\def\year{2018}\relax
\documentclass[letterpaper]{article} 
\usepackage{aaai18}  
\usepackage{times}  
\usepackage{helvet}  
\usepackage{courier}  
\usepackage{url}  
\usepackage{graphicx}  
\usepackage{algorithm}
\usepackage{algorithmic}
\usepackage[utf8]{inputenc}
\usepackage{amsmath,amsfonts,amssymb,bbm, amsthm}

\newcommand{\RN}[1]{%
	\textup{\uppercase\expandafter{\romannumeral#1}}%
}
\begin{document}
 	
%
\title{ Risk-Stratify: Confident Stratification Of Patients Based On Risk}
\author{Kartik Ahuja,  Mihaela van der Schaar
}
\maketitle
\begin{abstract}
Risk-stratification divides the population (patients at risk of some disease) into groups with different risks. The clinician desires to use a risk-stratification method that  achieves \textit{confident risk-stratification} - the risk estimates   of the different patients reflect the true risks  with a high probability. This allows him/her to use these risks to make accurate predictions about prognosis and decisions about screening, treatments for the current patient. We develop Risk-stratify - a two phase algorithm that is designed to achieve confident risk-stratification. In the first phase, we grow a tree to partition the covariate space.  Each node in the tree is split using statistical tests that determine if the risks of the child nodes are different or not. The choice of the statistical tests depends on whether the data is censored (Log-rank test) or not (U-test). The set of the leaves of the tree form a partition. The risk distribution of patients that belong to a leaf is different from the sibling leaf but not the rest of the leaves. Therefore, some of the leaves that have similar underlying risks are incorrectly specified to have different risks. In the second phase, we develop a novel recursive graph decomposition approach to address this problem. We merge the leaves of the tree that have similar risks to form new leaves that form the final output. We apply Risk-stratify on a cohort of patients (with no history of cardiovascular disease) from UK Biobank and assess their risk for cardiovascular disease. Risk-stratify significantly improves risk-stratification, i.e., a lower fraction of the groups have over/under estimated risks (measured in terms of false discovery rate; 33\% reduction) in comparison to state-of-the-art methods for cardiovascular prediction (Random forests, Cox model, etc.). We find that the Cox model significantly over estimates the risk of 21,621 patients out of 216,211 patients. Risk-stratify can accurately categorize 2,987 of these 21,621  patients as low-risk individuals.

\end{abstract}
\vspace{-1em}

\section{Introduction}

\subsection{Motivation} 
Risk-stratification divides the population (patients at risk of some disease) into groups with different risks (probability of developing a disease up to a certain time). A clinician desires to use risk-stratification methods to estimate the risks and arrive at reliable decisions  (screening actions, interventions, treatments).  For instance, the seminal work \cite{fonarow2005risk} develops a  risk-tree for mortality risk-stratification of patients hospitalized with acutely decompensated heart failure (ADHF). This risk-tree  is aimed to assist the clinician's decisions - low-risk patients  get treated less intensively in the telemetry ward, while  high-risk patients get treated in the intensive care unit. In cardiovascular disease (CVD),  the commonly used models for risk-stratification are  the Framingham risk score \cite{d2001validation}, and  the QRISK score \cite{lin2013ankle}. These models assist the clinicians with decisions regarding interventions such as anti-platelet therapy \cite{baigent2009aspirin} and cholesterol lowering medications (See, e.g.,  http://qintervention.org/ which uses QRISK score).  In breast cancer, the commonly used models for risk-stratification are the IBIS model \cite{tyrer2004breast}, and the Gail model \cite{gail1989projecting}.  These models are used for screening  \cite{alaa2016confidentcare} and  treatment recommendations such as Tamoxifen \cite{coombes2007survival}.  

The clinician expects that the risks estimated  based on these models for different patients (or at least groups of patients)   reflect the true risks accurately.  However, we find that these models can fail to do so. For instance,  the QRISK score \cite{lin2013ankle}, which uses a Cox proportional hazards model, overestimates the risks.  We used this model to estimate the risks of CVD for individuals in UK Biobank data set \cite{biobank2007uk} (details in Motivating Example and Experiments).  The model  significantly overestimated the risk of 21,621 patients. This can mislead the clinicians to prescribe treatments such as anti-platelet therapy, cholesterol lowering medications, which are known to have strong side-effects such as  excess bleeding \cite{baigent2009aspirin}, intestinal problems. Therefore, it is \textit{essential} to develop methods for accurate risk-stratification.

\subsection{Objective and Approach}
We develop  a method that stratifies the patients into groups and achieves \textit{confident risk-stratification}- the  order of risk estimates of the different groups is equal to the true order of the risks with a high probability.  We require the method to identify a minimum number of groups such that the clinicians can make decisions at a refined level.

We use false discovery rate (FDR) \cite{shaffer1995multiple} - a metric that measures the proportion of pairs  of groups that have very similar risks but are estimated to be very different. If the FDR is zero, then the order of  risk-estimates of different groups is completely accurate. The challenges in achieving confident risk-stratification are i) there is an exponentially large number of ways to group the patients and a finite amount of data; checking many possibilities simultaneously is prone to false discoveries  \cite{shaffer1995multiple}, ii) it is computationally intractable to achieve the exact solution. 

We developed Risk-stratify algorithm to address the above challenges. Risk-stratify constructs a partition of the covariate space to stratify the patients into groups. Risk-stratify consists of two phases. In the first phase, we  grow a tree to partition the covariate space. Each node in the tree is split along the covariate dimension that creates sufficiently different risk-distributions in the child nodes.  We do not make any parametric assumptions (such as proportional-hazards) thus we use non-parametric tests such as Log-rank test \cite{mantel1966evaluation}, and U-test  \cite{mann1947test}. The set of the leaves of the resulting tree form a partition. There are many leaves in the tree that have very similar risk distributions but are estimated to be very different. In the second phase of the algorithm, we address this problem. We develop a novel recursive graph decomposition approach that builds on ideas from graph-theory and hypothesis testing. The procedure merges  the leaves  that have very similar risk distributions to form a new partition that is close to risk-stratified.  Risk-stratify is very flexible regarding the types of regions (non-compact) that describe a partition. Therefore, it enables very effective risk-stratification. Risk-stratify provides interpretable partitions of the covariate space, which is desired for the implementation in  clinical practice. See Figure 1 for an overview of Risk-stratify.

Risk-stratify is applied on patients (with no history of CVD) in the UK Biobank data set to assess their risk for CVD. Risk-stratify significantly improves risk-stratification, i.e., a lower fraction of the groups have over/underestimated risks (measured in terms of FDR; 33\% reduction) in comparison to state-of-the-art methods for CVD prediction \cite{weng2017can}  such as random forests, Cox model, etc.). The Cox model  significantly overestimated the risk of 21,621 patients to medium-risk (15-42 percent higher than population risk); Risk-stratify correctly identifies 2,987  low-risk patients  out of these 21,621 patients.

\section{Problem Formulation and Methods}


\subsection{Preliminaries}
We consider a dataset $\mathcal{D}$ comprising  of survival (time-to-event) data for $N$ patients ($\mathcal{N}=\{1,..,N\}$ is the set of patients) who have been followed up for a finite amount of time. Let $\mathcal{D} = \{X_i, T_i\}_{i=1}^{N}$, where $X_i \in \mathcal{X}$ is a $d$-dimensional vector of covariates associated with patient $i$ and $T_i$ be  the observed failure time. Let $T_i^*$  be the true failure time, $C_i$ be the censoring time for patient $i$. The observed failure time is given as $T_i = \min\{T_i^{*}, C_i\}$.  For ease of exposition we assume each dimension of the covariate is a binary categorical variable. However, the method extends to non-categorical variables (See the Supplementary Material).  We define the survival distribution conditional on the covariates $X_i$ as $S(t|X_i) = Pr(T_i^*>t   |X_i)$.
We use  the Kaplan-Meier estimator \cite{kaplan1958nonparametric} for the survival function conditional on the covariates $X\in \mathcal{X}^{'}\subseteq \mathcal{X}$ and denote the estimate as $\hat{S}(.|\mathcal{X}^{'})$. A set $P=\{\mathcal{X}^{1},...,\mathcal{X}^{k}\}$  is a partition of $\mathcal{X}^{'}$ if it satisfies the following conditions i) any two regions $\mathcal{X}^{m}$ and $\mathcal{X}^{n}$ in the set $P$ do not intersect  $\mathcal{X}^{m}\cap \mathcal{X}^{n} = \Phi$, ii) $\mathcal{X}^{'} = \cup_{i=1}^{k} \mathcal{X}^{i}$.  Each patient $X$ belongs to  one of the groups $\mathcal{X}^{j}$ in the partition $P$.  Define $\mathcal{P} $ as the set of all the possible partitions of $\mathcal{X}$.

\subsection{Goal and Requirements:}

\textbf{Prognosis up to a given time:} The clinician wants to predict the risk - the probability of developing a disease up to a certain time $t^*$.  We formulate the risk-stratification problem as a search for the partition $P\in \mathcal{P}$ of the covariate space  such that the risk estimates for each region in the partition are close to the true risks and thus can be used by the clinician for prediction. We define the distance between the survival distribution at time $t^{*}$ in the two regions $\mathcal{X}^{m}$, $\mathcal{X}^{n}$  as
	\begin{equation}
	D(S(.|\mathcal{X}^{m}), S(.|\mathcal{X}^{n})) = |S(t^{*}|\mathcal{X}^{m}) -S(t^{*}|\mathcal{X}^{n})|
	\label{eqd1}
	\end{equation}

\textbf{Clinically significant separation:}   We say two regions $\mathcal{X}^{m}$ and $\mathcal{X}^{n}$ have a clinically significant separation if the difference in the risks \eqref{eqd1} is greater than some specified theshold $\Delta$. For instance, $\Delta= \frac{\overline{Risk}}{10}$, where $\overline{Risk}$ is the average population risk.

\textbf{Risk-Stratified  Partition:}   If in a partition $P$  every two distinct regions ($\mathcal{X}^{m} \in P, \mathcal{X}^{n}\in P$, $m\not=n$)  have a clinically significant separation, then the partition is risk-stratified. A risk-stratified partition $P$ can also be understood as a set of regions in the covariate space, which are ordered in terms of their risks (where risks differ by at least $\Delta$).


\textbf{Maximum Risk-Stratified  Partition:}  There can be many risk-stratified partitions. Given a risk-stratified partition $P=\{\mathcal{X}^{1},..,\mathcal{X}^{k}\}$ we can construct another risk-stratified partition $P^{'}=\{\mathcal{X}^{1}\cup \mathcal{X}^{2},\mathcal{X}^{3}..,\mathcal{X}^{k}\}$ from $P$. The partition $P$ is more desirable than than $P^{'}$ because it has more refined information. Therefore, we define the maximum risk-stratified partition $P^{*}$ to be the largest risk-stratified partition. 
\vspace{-0.5em}
\begin{equation}
\begin{split}
P^{*} = &\arg\max_{P} \;\;|P| \\
& \text{s.t.} \;\; 	P \;\text{is risk-stratified}\\
\end{split}
\label{eqn0}
\end{equation}
If there are ties in \eqref{eqn0}, we select a partition that has the largest minimum separation between all the pairs of regions. \eqref{eqn0} cannot be solved exactly because the underlying  survival distribution is not known. Our objective is to use the data $\mathcal{D}$ to arrive at a partition that is close to $P^{*}$. Since $\mathcal{D}$ is random the output is random as well. Therefore, the problem in \eqref{eqn0} is stated in terms of probabilistic constraints.

\textbf{Confidence level:}  The clinician states a confidence level  $1-\alpha$  and requires that the output is risk-stratified with probability at least  $1-\alpha$.

\vspace{-1em}
\begin{equation}
\begin{split}
\hat{P}^{*} =&\arg\max_{P} \;\;|P| \\
& \text{s.t.} \;\; 	P \;\text{is risk-stratified with high probability}\\
\end{split}
\label{eqnapp}
\end{equation}

We require that $\hat{P}^{*} $ is risk-stratified with probability at least $1-\alpha$ and it is close to $P^{*}$ (detailed description in the next Section). Suppose $\hat{P}^{*}=\{\mathcal{X}^{1},...,\mathcal{X}^{4}\}$ ($\Delta=1\%$, $\alpha=0.05$); the risk estimates  for developing the disease in 5 years are $\{2\%,4\%,6\%,8\%\}$. Since $\alpha=0.05$, in  $95 \%$ of the cases, the order of the risks discovered are accurate.
%
%

%

\textbf{Challenges in solving \eqref{eqnapp}:} There are a large number of possible partitions and a finite amount of data. Testing on many partitions is prone to false discoveries \cite{shaffer1995multiple}. Also, \eqref{eqnapp} is computationally intractable (See Supplementary Material). Next, we discuss a motivating example to show how standard models fail in achieving risk-stratification.


\textbf{A Motivating Example:}

We consider a cohort in the UK Biobank dataset (details in the Experiments Section), where the event is defined as the development of CVD. We fit a Cox proportional hazards model (used in QRISK score).  Consider the two risk groups that are identified by the Cox model: Low risk ($2.5-4\%$)  group, and Medium risk ($4-5\%$.). Consider two patients one from each risk group: Subject $1$ is a male, who is less than $52$ years old, with a moderate BMI (between $25$ to $28$), and systolic blood pressure (below $128$), with no diabetes, no hypertension, no lipid lowering drugs, no smoking history, Subject $2$ is a female, who is less than $52$ years old, with a moderate BMI (between $25$ to $28$), and systolic blood pressure (between $128$ to $144$), with no diabetes, no hypertension, no lipid lowering drugs, no smoking history. The Cox model assigns Subject $2$ to the Low-risk group  (risk score of $2.5 \%$) and Subject $1$ to the Medium-risk group (risk score of $4\%$). These risk scores are the output of the estimated models, but they may not reflect the true risks. We compare the risk-distributions of the two groups based on the survival times. Based on the hypothesis-test (U-test) that compares the two risk groups  it is found that there is no evidence of a difference between the two groups (in fact both groups have a risk of less than $2.5\%$). The individuals such as Subject 1 who were assigned to risk-group $4-5\%$ have an overestimated risk and this can mislead a clinician to recommend unnecessary treatments.

\subsection{Hypothesis-Testing Based Partition Search}

 We begin by describing a naive approach to solve  \eqref{eqnapp}.  We use the  Kaplan-Meier estimates of the survival distributions for every pair of regions to check for clinically significant separation.  There is a serious limitation of this approach. The empirical estimate of the distance can have a high variance. Therefore, it is likely to  identify many partitions that are not risk-stratified as risk-stratified. We propose a hypothesis-testing approach that overcomes the limitations of the naive approach. We define a null hypothesis and an alternate hypothesis to check  risk-stratification.

\begin{itemize}
\item $H_0(P)$: $P$ is not risk-stratified
\item $H_1(P)$: $P$ is risk-stratified
\end{itemize}	

%
%

We define a hypothesis test $\mathcal{T}(\mathcal{D}, \mathcal{P})$.  The test takes the entire data $\mathcal{D}$ and the possible set of partitions $\mathcal{P}$ as the input and classifies them as risk-stratified ($H_1(P)$ is true) or not ($H_0(P)$ is true).  The set of partitions that are classified as risk-stratified by the test is defined as $\mathcal{P}^{+}$.

  \textbf{\textit{False Discovery Rate:}}  Suppose there are multiple hypothesis that are  tested and some of them are declared  positive.  False discovery rate  \cite{shaffer1995multiple} is defined as the proportion of the false positives among the  hypotheses that were declared positive.   If a partition $P$ is tested for risk-stratification and is declared risk-stratified (positive), then it is equivalent to saying that $|P|(|P|-1)/2$  pairs of distinct regions in the partition are declared to have a clinically significant separation.    Suppose  $W$ of the $|P|(|P|-1)/2$  comparisons are false positives, then the $FDR(P)$ is  \begin{equation}
FDR(P) =  \frac{W}{|P|(|P|-1)/2} 
\label{FDR}
  \end{equation}  $FDR(P)$ is the fraction of pairs of regions in the partition that have similar risks but are estimated to be very different. 
  

    We reformulate the  problem in \eqref{eqnapp} in terms of constraints on the $FDR$. If $FDR(P)=0$, then the partition is risk-stratified. If $FDR(P)>0$, then the partition is not risk-stratified.  We need to ensure that the confidence level constraint from the clinician is met.  We restrict the probability that at least one partition, which is not risk-stratified,  is output by the test to be less than $\alpha$ and state the risk-stratification problem as

    \vspace{-1em}
  \begin{equation}
  \begin{split}
  & \tilde{P}=\;\;\;\;\;\;\;\;\arg \max_{P}\;\;\;\;\;\;\; |P|\\ 
  & \text{s.t.}\; \;\;\;\;\;\;\;\;\;Pr(\cup_{P \in \mathcal{P}^{+}} \{FDR(P)>0\}) \leq \alpha, P \in \mathcal{P}^{+}
  \end{split}
  \label{eqn4}
  \end{equation}

$\tilde{P}$ is a risk-stratified partition, i.e., the order of the estimated risks of the regions in $\tilde{P}$ reflect the true order of the risks, with a probability $1-\alpha$. 


We now propose a test $\mathcal{T}$ that satisfies the constraint described above in \eqref{eqn4}. 
 Consider any two regions $\mathcal{X}^m \in P$, $\mathcal{X}^n \in P$, $m\not=n$. The dataset $\mathcal{D}$ is divided into these regions; we denote the dataset in region $\mathcal{X}^m$, $\mathcal{X}^n$ as $\mathcal{D}^m$, $\mathcal{D}^n$ respectively.   We compare the distance $D$ between the survival distributions in these two regions $\mathcal{X}^m$, $\mathcal{X}^n$ using an appropriate hypothesis test.  If  all the patients have been followed upto time $t^{*}$, then we use the U-test, else we use log-rank test.  Define $Pv(\mathcal{X}^{m}, \mathcal{X}^{n}, \mathcal{D})$ to be the p-value based on the test that compares the risks in the  regions $\mathcal{X}^{m}$ and $ \mathcal{X}^{n}$. The set of partitions output  by the test $\mathcal{T}$  is given as

\vspace{-1em}
 $$\mathcal{P}^{+} = \{P : Pv(\mathcal{X}^m,\mathcal{X}^n,\mathcal{D}) \leq \alpha^{'}, \forall \mathcal{X}^m \in P, \mathcal{X}^n \in P, m \not=n\}$$
 
If $\alpha^{'}=\frac{\Gamma \alpha}{|P|(|P|-1)|\mathcal{P}|}$, then the test $\mathcal{T}$ described above satisfies the constraints in \eqref{eqn4} (See Supplementary Material for details and value of $\Gamma$). We restate \eqref{eqn4} in terms of the constraint on the p-values.
  

%
    
\vspace{-1em}
\begin{equation}
\begin{split}
& \bar{P}= \;\;\;\;\;\;\;\;\;\arg \max_{P } \;\;|P|\\ 
& \text{s.t.}\;\;\;\;\;Pv(\mathcal{X}^m,\mathcal{X}^n,\mathcal{D}) \leq \alpha^{'}, \forall \mathcal{X}^m , \mathcal{X}^n \in P, m\not=n 
\label{eqn6}
\end{split}
\end{equation}

There are a few problems associated with directly solving \eqref{eqn6}. The set $\mathcal{P}$, in general, is very large thus the threshold $\alpha^{'}=\frac{\Gamma \alpha}{|P|(|P|-1)|\mathcal{P}|}$ can be very restrictive (limiting the set of partitions that satisfy the constraint).  \eqref{eqn6} is computationally intractable (See  Supplementary Material). 
To address the concerns about \eqref{eqn6}, we develop a greedy approach to solve the above optimization problem. 

\begin{figure}
	\begin{center}
		\includegraphics[trim= 35mm 50mm 0mm 0mm, width=4.5in]{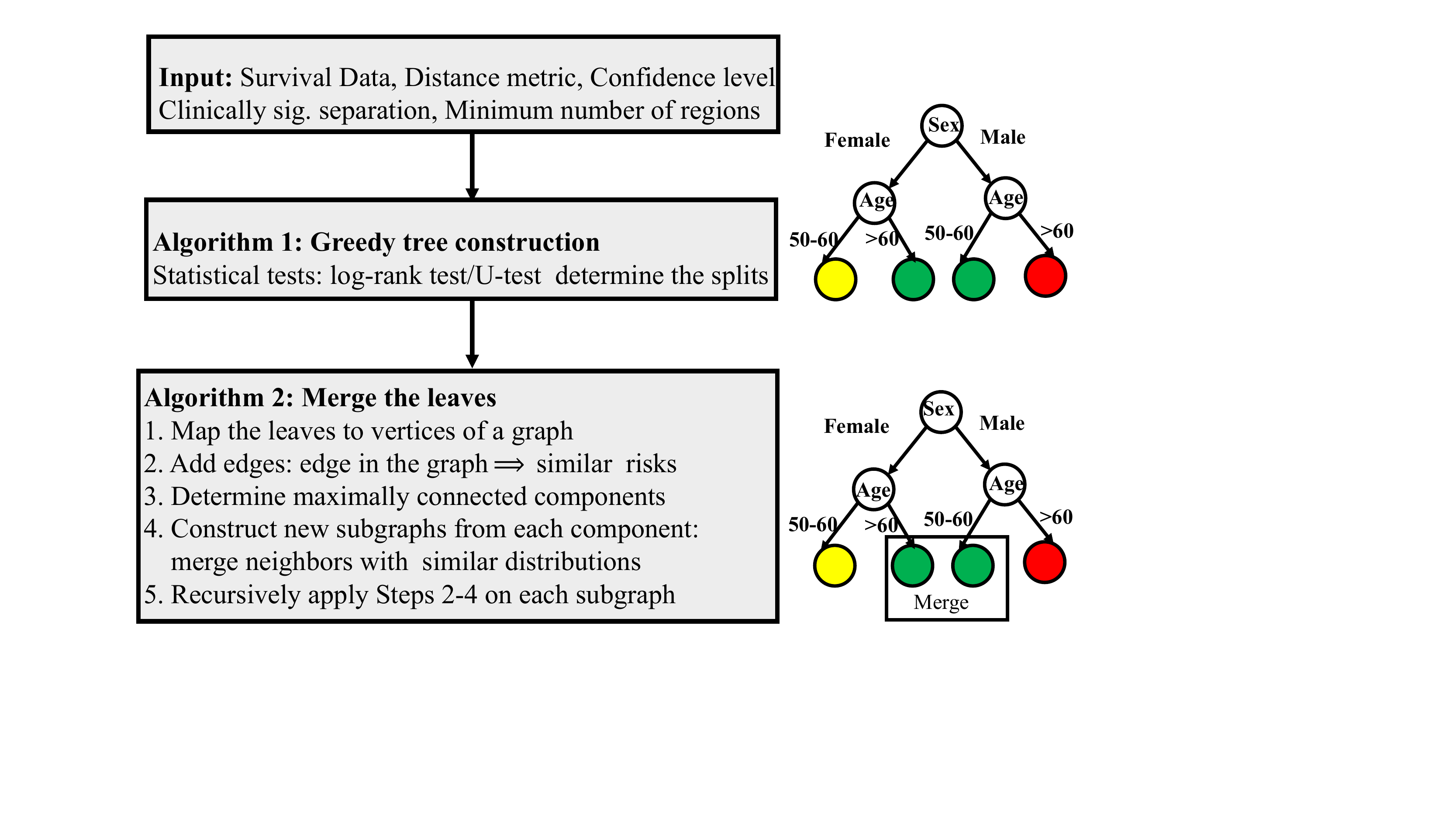}
		\caption{Overview of Risk-Stratify}
		\label{Illustration2}
	\end{center}
\end{figure}

\subsection{Methods: Risk-Stratify}

In this section, we describe the Risk-stratify method. The method consists of two phases: Algorithm 1- Greedy Tree Construction and Algorithm 2- Merge the  Leaves.
\subsubsection{Algorithm 1- Greedy Tree Construction:}

We define a function $\Theta$ that takes as input the covariate space $\mathcal{X}$ and outputs a partition $\Theta(\mathcal{X})$ described as
\begin{equation}
\begin{split}
\Theta(\mathcal{X}) =
& \arg \max_{P}\;\;\;    |P| \\
& \text{s.t.} \;\; Pv(\mathcal{X}^m,\mathcal{X}^n,\mathcal{D}) \leq \alpha^{'}, \forall \mathcal{X}^m , \mathcal{X}^n \in P, m\not=n \\
&\;\;\;\;\;\; |P| \leq 2,\; P \;\text{is composed of hypercubes} 
\label{eqn5}
\end{split}
\end{equation}

In \eqref{eqn5}, the optimization consists of two more constraints added to \eqref{eqn6}. The first constraint restricts the size of the partition to be less than or equal to two and the second constraint restricts the partition to be only composed of hypercubes, where a hypercube is a set that can be written as $[a_1,b_1] \times... [a_d, b_d]$. If there are ties, then we choose a partition that minimizes the maximum of the p-values  across every pair of regions in the partition.

We use the above function \eqref{eqn5} to develop the greedy approach in Algorithm 1.
The main idea goes as follows: divide the covariate space into  two regions (if feasible) and then apply the function in \eqref{eqn5} on the generated regions in the partition. 
We describe the pseudo-code for the above procedure in Algorithm \ref{Algo1}. The output of Algorithm 1 is $\mathcal{L}$.

  \begin{algorithm}[tb]
	\caption{ Risk-stratify: Greedy Tree Construction}
	\label{alg:example}
	\begin{algorithmic}
		\STATE Fix the hyperparameter $\alpha=0.05$, $\alpha^{'}=\frac{\alpha}{d}$ 
		\STATE \textbf{Input:} $\mathcal{X}$
		\STATE \textbf{Output:} $\mathcal{L}$ = GreedyOpt($\mathcal{X}$)
		\STATE \hrulefill
		\STATE \textbf{\textit{Function}} : GreedyOpt($\mathcal{Z}$) 
		\STATE $\mathcal{Z}^{'}=\Theta(\mathcal{Z})$
		\STATE $\mathcal{M}=\Phi$
		\STATE  \textbf{If} ($\mathcal{Z}^{'}==\mathcal{Z}$)
		\STATE $\;\;\;\;$   $\mathcal{M}=\mathcal{Z}$
		\STATE \textbf{else}
		\STATE $\;\;\;\;$\textbf{for} $\mathcal{Z}_j \in \mathcal{Z}$
		\STATE $\;\;\;\;\;\; \mathcal{M} = \mathcal{M} \cup \text{GreedyOpt}(\mathcal{Z}_j)$
		\STATE $\;\;\;\;$\textbf{end for}
		\STATE \textbf{end if}
		\STATE \textbf{\textit{Function output}:} $\mathcal{M}$
	\end{algorithmic}
\label{Algo1}
\end{algorithm}

\textbf{Comments on Algorithm 1:}  The worst case computational complexity of the Algorithm \ref{Algo1} is $\mathcal{O}(N^3d)$ (See Supplementary Material for the proof), where  $N$ is  the number of patients and $d$ is the length of the covariate vector.  We   control  the variance in the risk-estimates in each region of the partition by including an explicit constraint on the minimum number of points in the region (See Experiments). 



\textbf{Limitations of Algorithm 1:} In Algorithm 1, at each split, we ensured that the two child nodes/regions that are generated have sufficiently different risk distributions. However,  there are many leaves that have similar true risks but are estimated to be very different. In Algorithm 2, we  address this issue. In Algorithm 2, we   merge the leaves that were estimated to have very different risks (with similar true risks) in such a way that the new merged leaves form a risk-stratified partition and at the same time still allow the partition to have a large size. In the next Section, we describe Algorithm 2 (pseudo-code in the Supplementary Material).

%
%

%

\textbf{Algorithm 2- Merge  the Leaves:}
 In Algorithm 2, we build a recursive graph decomposition approach to merge the leaves.   We provide a target number of leaves $Nleaf$  as a termination criterion. 
We summarize the key steps and the rationale for the Steps of the Algorithm 2 below.

\begin{enumerate}
	\item  Suppose the first phase of the algorithm outputs a partition   with $L$ regions (leaves). We create an empty graph $G$ with $L$ vertices, where each vertex corresponds to a leaf in the tree.  For each vertex add an attribute called data, which contains the survival data for  the patients in the leaf.
	\item For every pair of vertices in the graph $G$ do a hypothesis test (Log-rank test / U-test) and compute the p-values associated with each pair. If a pair has a p-value higher than the threshold $\alpha^{'}$, then connect it by an edge. This step identifies all the pairs of regions/leaves output from Algorithm 1 that share similar distributions (violate the risk-stratification constraint). There can be many possible ways to merge the vertices connected by the edges. We adopt a principled approach based on the statistical significance of the hypothesis tests.
	\item   Compute all the maximally connected components of the graph $G$ using  standard breadth first search method  (See \cite{hopcroft1973algorithm}). Consider two vertices in two different maximally connected components. These vertices  have sufficiently different risks, and thus there is no need to merge the vertices in two different components. Hence, each component should be dealt with separately as described in the next step. 
	\item  We use each connected component identified in Step 3 to create a new subgraph described as follows. We use the p-values to identify the nearest neighbors in terms of risks. For each vertex $v$ in the  connected component  identify the edge $(v,w)$ associated with the highest p-value.  If the vertex $w$ has not been already merged with another vertex, then merge vertex $v$ and $w$ to form a new vertex for the new  subgraph. If the vertex $w$ has already been merged to form a vertex $v^{'}$, then add the vertex $v$ to the new vertex $v^{'}$. Also, add the survival data associated with the vertices that are merged to the data attribute of each new vertex.
	\item Recursively apply the Steps 2-4 on every  new subgraph (if it is not a singleton)   until  the total number of vertices (across all the subgraphs) equals $Nleaf$.
\end{enumerate}

\textbf{Comments on Algorithm 2:}  The worst case computational complexity of Algorithm grows as $\mathcal{O}(N^3)$ (See the Supplementary Material for the proof). We fixed the hyperparameter $\alpha^{'}$ used in Algorithm 1 and 2 to $\frac{0.05}{d}$ (See justification in Supplementary Material). Next, we provide an illustrative example to highlight how Risk-stratify works and the advantages offered by it in comparison to other methods. 

%
%

\begin{figure}
	\begin{center}
		\includegraphics[trim= 0mm 30mm 0mm 0mm, width=3.7in]{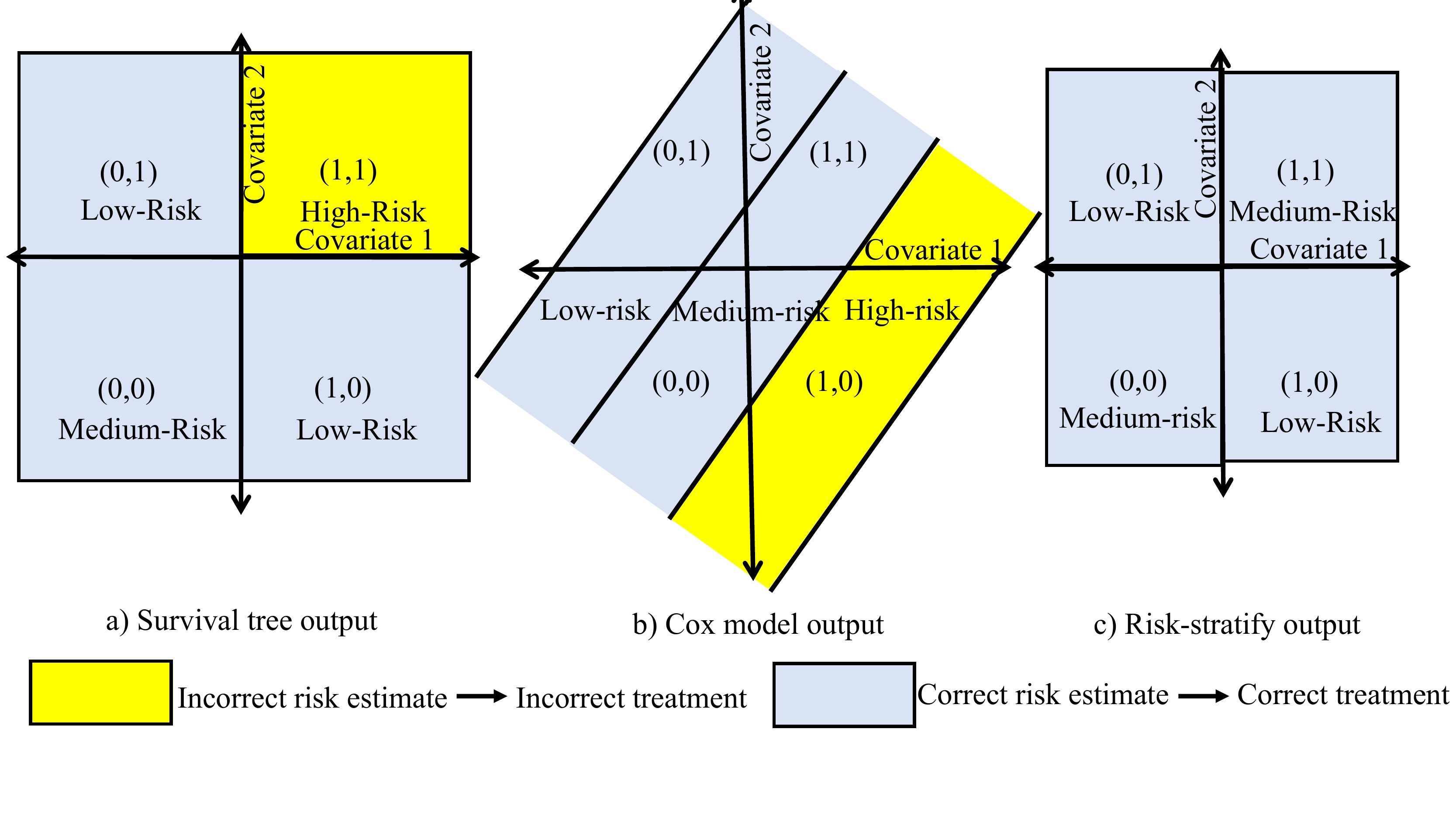}
		\caption{Comparison of regions (color coded) created for risk-stratification using different methods}
		\label{Illustration2}
	\end{center}
\end{figure}

\textbf{Illustrative Example:}

Consider a survival data set, where each patient has two covariates $x_1,x_2$ ($x_1$ indicates whether the patient is obese or not, $x_2$ indicates if the patient had menopause or not).  The clinician is interested in stratifying the probability of developing cancer in $5$ years and requires to identify (at least) $2$ risk-stratified groups  with a separation of $\Delta=5\%$. Define three risk groups low-risk $0-10 \%$, medium-risk $10-20\%$,  high-risk $>20 \%$. We assume  the survival distributions (based on \cite{lophatananon2017development}) are 
\begin{equation}
S(t|(x_1,x_2)) = 
\begin{cases}
\frac{5}{1+t} \;\; \;\text{if} \;x_1 \oplus x_2 = 0  \\
\frac{10}{1+2t} \;\; \text{if}\; x_1 \oplus x_2 = 1 
\end{cases}
\end{equation}



where  $\oplus$ is the XOR operation. 
For the above model we know the exact solution to \eqref{eqn0}, $(0,0)$ and $(1,1)$ should be in medium-risk group since the true risk is $16\%$ and $(0,1)$ and $(1,0)$ should be in low-risk group since the true risk is $9\%$. 
We simulate the data from the above model. In Figure \ref{Illustration2}, we plot the 2-dimensional feature space. Cox models use linear functions to model the impact of covariates on the hazard rate. The Cox model divides the covariate space into three regions, where $(0,1)$ is low-risk , $(0,0)$ and $(1,1)$ as medium-risk, and $(1,0)$ as high-risk. Cox model overestimates the risk of $(1,0)$. In this case  $FDR= 1/3$.

Survival tree models \cite{bou2011review} construct partitions where each region of the partition is a hypercube. We grow a survival tree  with four leaves. The risk estimates output from a survival tree for each of the leaves  are $(0,0)$- Medium-risk, $(1,1)$-  High-risk , $(1,0)$ - Low-risk , $(0,1)$ - Low-risk.  Survival tree overestimates the risk of $(1,1)$. In this case the $FDR=1/3$.


Algorithm 1 in Risk-stratify outputs the same four regions as described by the survival tree. Algorithm 2 in Risk-stratify merges the regions that are similar in survival distributions.  Risk-stratify arrives at the correct risk-stratified groups namely: $(0,0)$ and $(1,1)$ into the low-risk group, $(1,0)$ and $(0,1)$ into the group medium-risk group. In this case, $FDR=0$. 
This example illustrates that Risk-stratify allows complex regions (non-compact) to represent the partition. 
Both Cox model and Survival tree overestimate the risks and can mislead the clinician to prescribe unnecessary treatments such as Tamoxifen.

Before we describe the experiments we give another example of a distance metric $D$ where stratification is useful.

 \textbf{Risk-stratification of the entire survival distributions:}  The clinician wants to estimate the entire survival distribution. In this case, use the distance between the entire survival distributions in the two regions  $\mathcal{X}^{m}$, $\mathcal{X}^{n}$  defined as
\begin{equation}D\big(S(.|\mathcal{X}^{m}), S(.|\mathcal{X}^{n})\big)=\int_{0}^{\infty}|S(\tau|\mathcal{X}^{m})- S(\tau|\mathcal{X}^{n})|d\tau
\label{eqd2}
\end{equation}
In the above setting, we use Risk-stratify with log-rank test.
\vspace{-1em}
\section{Experiments}

In this section, we  describe the experiments conducted (in the R programming language) to evaluate Risk-stratify.

\textbf{Dataset:}  UK Biobank recruited half a million participants aging 40-69  between 2006-2010. The patients were invited for an initial assessment visit.  The health records for these patients (Hospital Episodes) were linked with UK Biobank. In our experiments, the event is the development of Cardiovascular Disease (CVD). We use the covariates that have been used in the clinical literature \cite{hippisley2007derivation} and \cite{weng2017can} for CVD prediction. The covariates that we use are: i) Sex, ii) Age, iii) Body Mass Index (BMI), iv) Systolic Blood Pressure, v) History of Diabetes, vi) Smoking Status, vii) Lipid lowering drug status, viii) Anti-hypertensive drug status.  We include all the patients who have no history of CVD. The total number of patients that are included are $432,225$ and  $3.5\%$ of these patients have a CVD event. 

\textbf{Data Preprocessing:}  We convert the continuous variables (age, blood pressure, BMI) into discrete variables, which are encoded as categorical variables. Divide each continuous variable value into three groups - Low, Medium, High, where Low is $<33 $ percentile values,  Medium is $33-66$  percentile values, and High is $>66$ percentile. 

\textbf{Requirements:} We  evaluate  Risk-stratify  in two different evaluation setups. 

 \textbf{Evaluation Setup 1:} The distance is measured between the survival probability at the time of interest (as defined in \eqref{eqd1}), where $t^{*}=5$ years. For every patient we know if they develop CVD at five years or not (no effect of censoring).

\textbf{ Evaluation Setup 2:} The distance is measured between the entire survival-distributions \eqref{eqd2} (impacted by censoring). 
 
  The population is at an average (5 years) risk of $3.5\%$ and  clinician specifies  $\Delta= 0.5\%$, which  is a reasonable choice (a group with risk $4\%$ is at a 15 percent higher risk than average risk). We set the minimum number of regions in the partition $Nleaf=10$.  In the evaluation setup 1 and 2, we use U-test and log-rank test respectively. We set the confidence level $1-\alpha=0.95$.

\textbf{Metrics and Cross-Validation based Evaluation:} 
There is a trade-off between the  FDR and the size of the partition (See \eqref{eqn4}). We compare the FDR across different methods by fixing the size of the partition to be the same across different methods. We divide the data randomly into two equal splits.\footnote{We use equal splits and not highly unbalanced splits because otherwise out-of-sample data may not even have sufficient values or any values at all in the partition identified.} We run the methods on the first half and compute the partition.  Every method outputs a partition, and our goal is to compute the FDR defined in \eqref{FDR} on out of sample data. The numerator, i.e. the true positives in FDR \eqref{FDR} is not known. Therefore, we estimate the out of sample FDR. We do a pairwise comparison between every region in the partition that is output by a method. If the p-value is below the adjusted significance level (adjustment based on Bonferroni correction  \cite{shaffer1995multiple}), then the comparison is declared to be a true positive, else it is declared to be a false positive.  We repeat the cross-validation procedure and compute the average of the FDR estimates across ten different  runs. In the Supplementary Material, we establish a relationship between the estimate of FDR computed using this procedure and the true FDR in \eqref{FDR} to justify our choice for estimating FDR. 



%
 
\begin{center}
	\begin{table}
		\caption{Comparing Risk-stratify for stratifiying survival probabilities at 5 years. $ 0.01<p \leq 0.05^{*},\;  p\leq 0.01^{**}$.}
		\begin{tabular}{ | l | c | r|}	
			\hline
			Method & FDR $\% \pm 95 \%$ CI  &  $\#$ of groups \\ \hline
			Risk-stratify & 8.7 $\pm$ 2.8 & 13.3 \\ \hline
			CART & 12.9 $\pm$ 5.1$^{*}$ & 13.3 \\ \hline
			Linear Classifier &13.4 $\pm$ 1.1$^{**}$  & 13.3 \\ \hline
			Logistic Classifier & 12.7 $\pm$ 1.5$^{*}$ & 13.3\\  \hline
			XGBoost & 13.3 $\pm$ 2.0$^{*}$ & 13.3\\ \hline
			Random Forest & 14.1 $\pm$ 1.5$^{**}$&13.3  \\
			\hline
		\end{tabular}
	\end{table}
\end{center}

\vspace{-2em}
\textbf{Benchmarks for Evaluation Setup 1-risk-stratification for survival probabilities at 5 years}
Classification and regression tree (CART) are natural candidates for comparison with Risk-stratify for Evaluation Setup 1 (no impact of censoring). We build the classification tree (RPART package) and we control the number of leaves to be equal to the number of regions output by Risk-stratify. We also implemented standard linear and logistic regression (glm function) to have a comparison with standard baselines. We also compare with random forest (randomforest package) and xgboost (xgboost package), which are the state-of-the-art machine learning methods for CVD prediction \cite{weng2017can}. We used standard settings for the hyperparameters of these methods. We create partitions for the linear, logistic, random forest, and xgboost models as follows. Fix the number of regions in the partition to be the same as in Risk-stratify. Divide the cumulative distribution of the risk scores that are output from these models into regions (low to high risk) such that every region has the same number of patients. (Further details in Supplementary Material)


\textbf{Benchmarks for Evaluation Setup 2- stratification of the  entire survival distributions} Survival trees are the natural candidates to compare with Risk-stratify for Evaluation Setup 2 (censoring present). We build the survival tree (RPART package) and we control the number of leaves  to be equal to the number of regions output by Risk-stratify. We also compare with proportional hazards survival models-Cox model and Weibull regression (survival package). These models provide a natural way to stratify the survival distributions, which is not true of non-proportional hazards model.  We use the proportional hazards term that depends on the covariates to construct the partitions (we use the procedure described for linear, logistic, randomforest and xgboost in the previous paragraph).
\begin{figure}
	\begin{center}
		\includegraphics[trim= 0mm 0mm 0mm 0mm, width=3.5in]{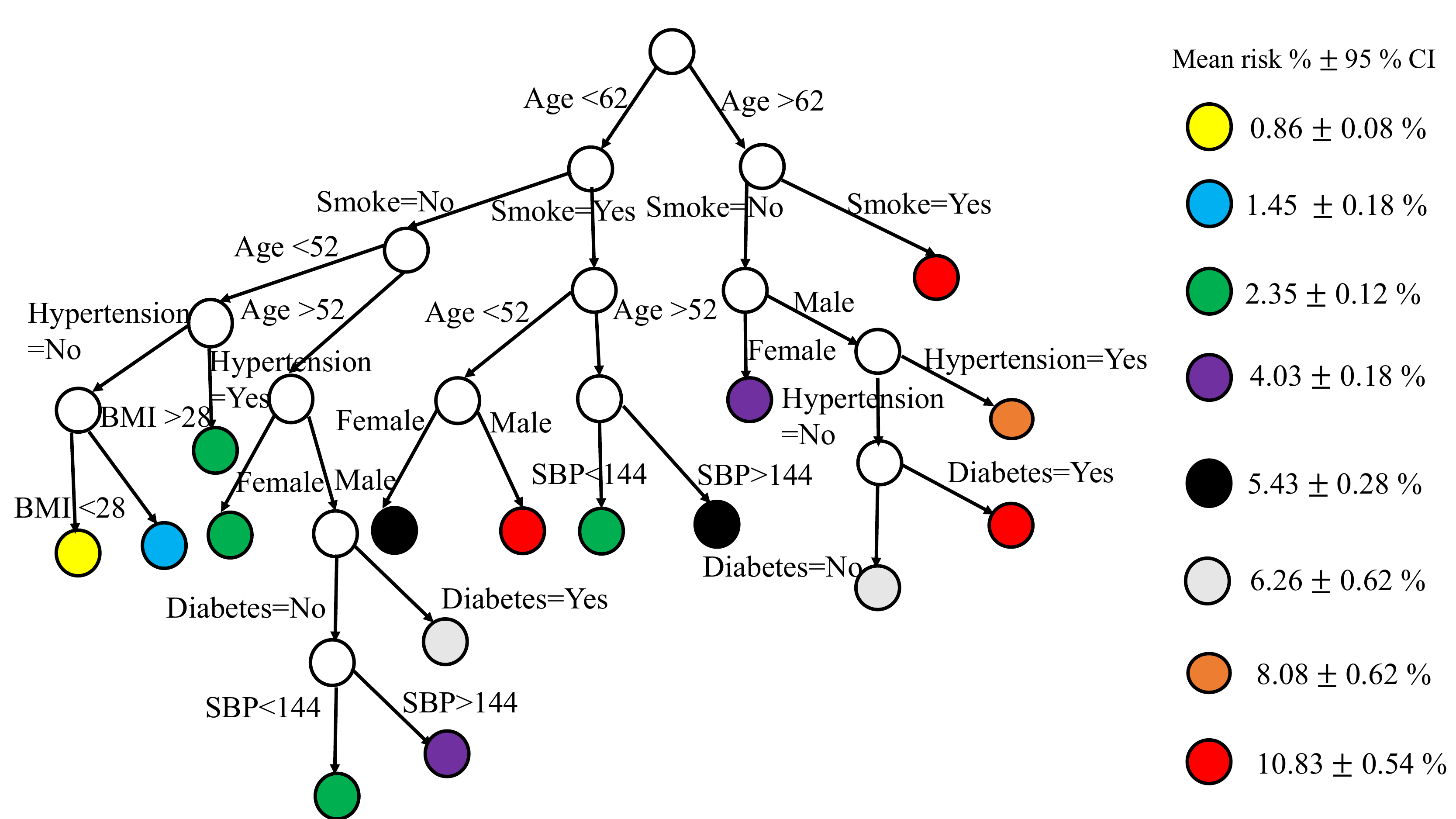}
		\caption{Risk stratification tree from Risk-stratify}
		\label{Illustration5}
	\end{center}
\end{figure}

\begin{figure}
	\begin{center}
		\includegraphics[trim= 0mm 20mm 0mm 0mm, width=2.7in]{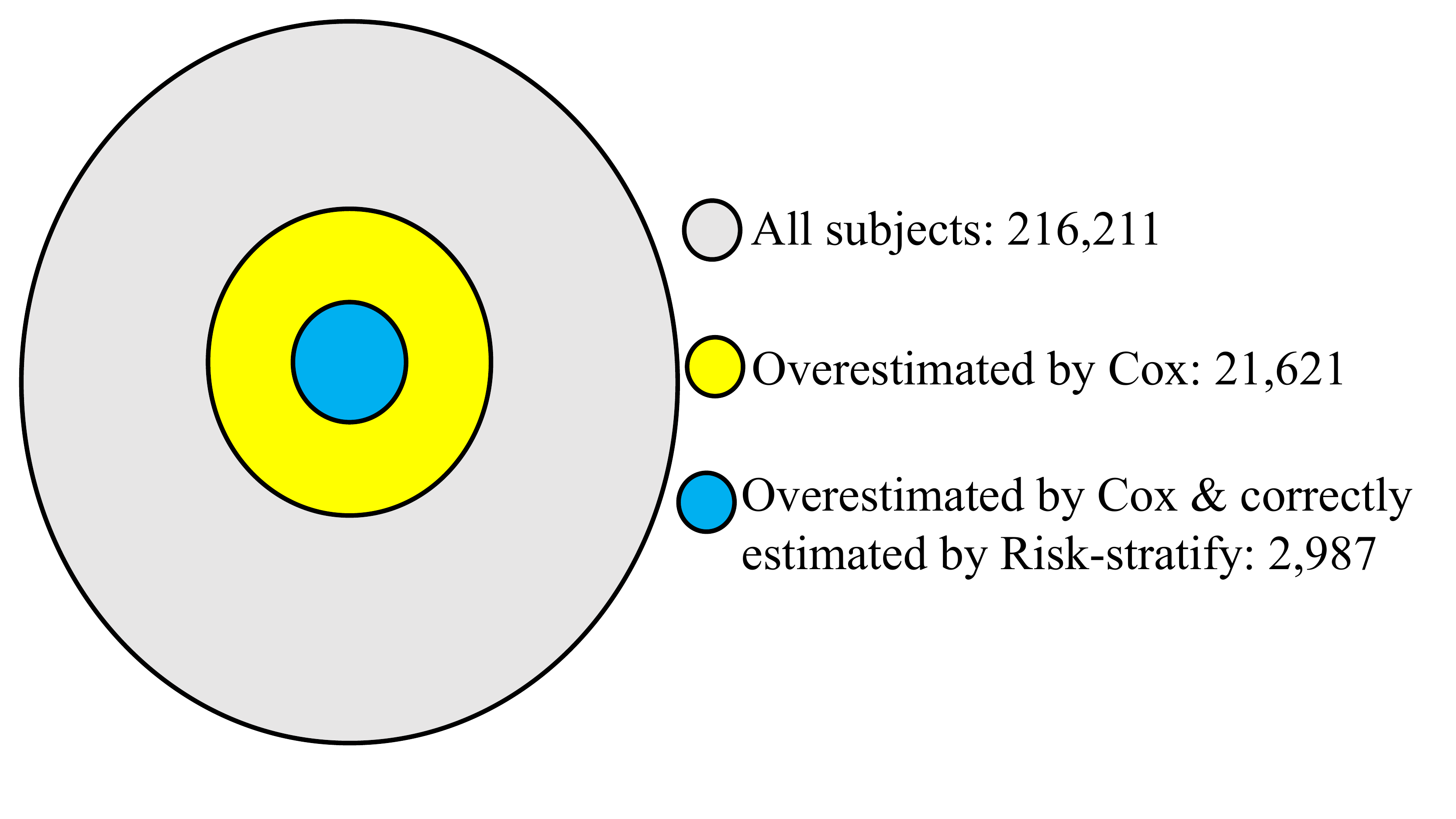}
		\caption{Illustrating risk-stratification gains}
		\label{Illustration7}
	\end{center}
\end{figure}

\begin{center}
	\begin{table}
		\caption{Comparing Risk-stratify for stratification of entire survival distribution. $ 0.01<p \leq 0.05^{*},\;  p\leq 0.01^{**}$.}
		\begin{tabular}{ | l | c | r|}	
			\hline
			Method & FDR $\% \pm 95 \%$ CI  & $\#$ of  groups \\ \hline
			Risk-stratify & 15.1 $\pm$ 1.9 & 17.6 \\ \hline
			Cox Regression & 18.1 $\pm$ 1.6 $^{*}$& 17.6 \\ \hline
			Weibull Regression &18.1 $\pm$ 1.4$^{**}$  & 17.6 \\ \hline
			Survival Trees & 22.6 $\pm$ 2.4$^{**}$ & 17.6\\  \hline
		\end{tabular}
	\end{table}
\end{center}
\vspace{-2em}
\textbf{Comparisons:} In Table 1, we show the comparisons for the Evaluation Setup 1 with distance in \eqref{eqd1}. Risk-stratify achieves an FDR of 8.7\%. The second best method logistic classifier has an FDR 12.7\% (33\% reduction).
In Table 2, we show the comparisons for the Evaluation Setup 2 with distance in \eqref{eqd1}. Risk-stratify  achieves an FDR of 15.1\%. The second best method Cox regression has an FDR 18.1\% (16\%  reduction). 
In  Table 1 and 2, we carry out t-test to compare whether the gain in the FDR is statistically significant or not (p-values are encoded as asterisks in Tables 1 and 2).  

\textbf{Interpreting the FDR gains:} The Cox model (second best method in Table 2) overestimates the risks of many patients. We find a group of  $21,621$ patients out of $216,211$ patients, which were assigned a high risk-score of $4-5\%$  (15-42 \% higher than the average population risk), whose true risks are $<4\%$.   Risk-stratify correctly estimates the risk of  $2,987$ patients out of these $21,621$ patients (These patients include Subject 1 from the motivating example). These $2,987$ patients that were significantly overestimated (given a risk of $4\%$) by the Cox model (while their true risks are $< 2.5\%$).  See Figure \ref{Illustration7}.


\textbf{Visualization and Interpretation} In Figure \ref{Illustration5}, we show the tree output by Risk-stratify. The merged leaves (with similar risks) are shown in the same color. Risk-stratify partitions the covariate space into interpretable regions unlike other methods (Details in Supplementary Material).

\vspace{-1em}

\section{Related Works}

%
%
\vspace{-0.5em}
\subsection{Survival Models}

\vspace{-0.25em}

\textbf{Non-Parametric Models-} \textbf{i) Censored Data:}
Survival tree based methods \cite{bou2011review} grow the tree based on statistical tests such as the log-rank test.  The leaves of the tree partition the entire covariate space.   The partition that is output from the survival trees does not satisfy the risk-stratification constraint. Each leaf represents a region with risks that are different from the sibling leaf but not the rest of the leaves. Therefore, some of the leaves that have similar underlying risks are incorrectly specified to have different risks.  Survival trees are similar to Algorithm 1 in Risk-Stratify.   These works differ from ours because we have a procedure to merge the leaves of the tree to ensure risk-stratification of the proposed partition. See Table \ref{table1} for a comparison of Risk-stratify with survival trees.

Ensemble based methods such as survival forests \cite{ishwaran2008random},  Bayesian additive regression trees \cite{chipman2010bart} aggregate survival trees. These methods do not output a partition of covariate space  that stratifies the survival distributions. 




\textbf{ii) Uncensored Data:}
If the survival data is uncensored, then classification and regression trees are useful for risk-stratification  \cite{breiman1984classification} \cite{loh2011classification}. These methods are similar to survival trees and share  similar limitations. The main difference between CART and survival trees is in the objectives used for tree growth such as Gini impurity in comparison to test statistic. See Table \ref{table1} for a comparison of CART with Risk-Stratify. Ensemble based methods such as random forests, boosting methods can also be used to stratify the patients (See Experiments Section). However, these methods cannot achieve effective stratification because these methods optimize loss functions such as accuracy, Gini impurity that does not necessarily guarantee that the risk estimates for different groups are accurate (further details in Supplementary Material). Moreover, the partitions that are output from these methods are not interpretable.

\begin{center}
	\begin{table}
		\caption{Comparing Risk-stratify, Survival Trees and CART}
		\begin{tabular}{ | c | c|}	
			\hline
			\textbf{Survival Trees \& (CART)} & \textbf{Risk-stratify}\\ \hline
	 		\textbf{Goal} & \textbf{Goal} \\ 
		Partition the population& Partition the population \\ 
	  into homogenous groups& into risk-stratified groups	 \\ 
		 		\textbf{Method} & \textbf{Method} \\ 
		 Recursive  partitioning & 	 Recursive partitioning\\ 
	& Algorithm 1\\ 
			 		\textbf{Metrics} & \textbf{Metrics} \\ 
	Test statistic (Info gain, Gini)& 		Constraint on p-values\\ \hline
		  		\textbf{Goal} & \textbf{Goal} \\ 
		  Avoid overfitting& Risk-stratification	 \\ 
	  		\textbf{Method} & \textbf{Method} \\ 
			  Cut branches & Merge leaves Algorithm 2	 \\ 
		    &   \\            
		  \textbf{Metrics}  & \textbf{Metrics}   \\      
  Size of leaf nodes& 	Target size, p-values	 \\ 
			  Size of the branch  &   Size of leaf nodes \\ \hline 
		\end{tabular}
		\label{table1}
	\end{table}
\end{center}

\vspace{-2em}
\subsubsection{Semi-Parametric and Parametric models:}
Cox proportional hazards models and other parametric proportional hazards models such as Weibull survival regression are the most commonly used models in survival analysis. These models do not output a partition of the survival distributions. These models use the proportional hazards assumption, which allows for the stratification of the survival distributions that is otherwise not possible in non-proportional hazards models.  These models make parametric assumptions that limit their stratification ability (As shown in the Illustrative Example and Experiments Section).

Recently, there is a large body  of work on phenotyping (See \cite{saria2015subtyping}, \cite{kim2017discriminative}).  Probabilistic mixture models \cite{schulam2015clustering} \cite{chen2016maximum} are commonly used in phenotyping and have some semblance to our work (We compare with other approaches to phenotyping such as tensor factorization in the Supplementary Material).   These probabilistic mixture models have also been used for survival analysis  (See \cite{ruhi2015mixture}, \cite{liverani2015premium}). The mixture based survival models are parametric and require distributional assumptions on the survival time and  the link functions. These assumptions limit these models in their ability to stratify.  Also, the above models, do not guarantee that patients in different mixture components have sufficiently different risks.

\vspace{-1em}
\section{Conclusion}

We developed Risk-stratify a novel method for risk-stratification.  The method takes as inputs - the confidence level associated with risk stratification and the minimum number of groups to be created.  Risk-stratify has two phases. The first phase uses a recursive partitioning approach (based on statistical tests) to construct a tree. In the second phase, we  use a novel recursive graph decomposition approach to merge the leaves of the tree to ensure the risk-stratification constraints are met.  We conduct experiments on UK Biobank dataset and show that the Risk-stratify  achieves significantly better risk stratification ($33\%$ reduction in false discovery rate, $2987$ patients correctly risk-stratified out of $21,621$ patients overestimated by Cox).

\section{Acknowledgement}

We would like to acknowledge Jinsung Yoon for his comments.
%

\bibliography{Tree_based_hypothesis}
\bibliographystyle{aaai}

\end{document}